\pdfoutput=1

\documentclass[11pt]{article}

\usepackage[final]{acl}
\usepackage{amsmath}
\usepackage{todonotes}
\usepackage{tikz}
\usepackage{pgf-umlsd}
\usepackage{listings}
\usepgflibrary{arrows} 
\usepackage{xcolor}
\usetikzlibrary{arrows,shapes,positioning,shadows,trees}

\usepackage{times}
\usepackage{latexsym}

\usepackage[T1]{fontenc}

\usepackage[utf8]{inputenc}

\usepackage{microtype}

\usepackage{inconsolata}

\usepackage{graphicx}
\usepackage{amssymb}
\usepackage{fontawesome5}  

\title{LiteWebAgent: The Open-Source Suite for VLM-Based Web-Agent Applications}

\author{Danqing Zhang$^1$ , Balaji Rama$^{1,2}$, Jingyi Ni$^1$, Shiying He$^1$, Fu Zhao$^1$ \\
 Kunyu Chen$^1$, Arnold Chen$^1$, Junyu Cao\thanks{~~~Corresponding author}$^3$\\
  $^1$PathOnAI.org\thanks{PathOnAI.org is an open-source AI research community on GitHub: \url{https://github.com/PathOnAI}}, 
$^2$Rutgers University, NJ, USA 
  $^3$The University of Texas at Austin, TX, USA \\
  \texttt{danqing.zhang.personal@gmail.com}, \texttt{balaji.rama@rutgers.edu}\\\texttt{junyu.cao@mccombs.utexas.edu} \\
  \href{https://www.pathonai.org/projects/litewebagent}{\faGlobe \hspace{1mm} https://www.pathonai.org/projects/litewebagent} \\
  \href{https://github.com/PathOnAI/LiteWebAgent}{\faGithub \hspace{1mm} https://github.com/PathOnAI/LiteWebAgent} \\
  \href{https://www.youtube.com/watch?v=lZUDbv5ABkg}{\faYoutube \hspace{1mm} https://www.youtube.com/watch?v=lZUDbv5ABkg}
}

\begin{document}
\maketitle

\begin{abstract}
We introduce LiteWebAgent, an open-source suite for VLM-based web agent applications. LiteWebAgent addresses a critical gap in the web agent ecosystem by providing an extensible core agent framework featuring planning, memory, and tree search capabilities, alongside a production-ready solution that combines minimal serverless backend configuration and intuitive user and browser interfaces. For the core LiteWebAgent agent framework, we implemented a simple yet effective baseline using recursive function calling, providing decoupled action generation and action grounding. In addition, we integrate advanced research components such as agent planning, agent workflow memory, and tree search in a modular and extensible manner. We then integrate the LiteWebAgent agent framework with frontend and backend systems in two deployment formats: (1) a production Vercel-based web application that provides users with an agent-controlled remote browser, and (2) a Chrome extension that leverages LiteWebAgent's API to control an existing Chrome browser via CDP (Chrome DevTools Protocol). The core LiteWebAgent framework is available at \url{https://github.com/PathOnAI/LiteWebAgent}, with deployed frontend at \url{https://lite-web-agent.vercel.app/}.
\end{abstract}
\section{Introduction and Related Work}

Recent advancements in large language models (LLMs) and vision-language models (VLMs) have revolutionized web browser automation. By late 2023, models like GPT-4V demonstrated the ability to handle real-world tasks on complex platforms such as Reddit and GitLab \cite{zhenggpt, zhou2023webarena}. This marked a shift from basic UI commands on toy web pages \cite{shi2017world} to sophisticated automation, driving research into VLM-powered agents for browser and device control. Companies like OpenAI, Anthropic, Adept, Google, and Apple are actively developing autonomous agents, recognizing their transformative potential.

While significant progress has been made, automation tasks can generally be categorized into two main approaches: web agents and device control agents. The following section outlines their differences and respective strengths.

\subsection{Web Agents vs. Device Control Agents}

Web agents and device control agents differ in their automation approaches. Web agents traditionally operate by parsing website structure - using DOM elements, HTML content, and accessibility trees - to enable LLM-based action generation and grounding. Recent advances have enhanced web agents by incorporating visual processing through Visual Language Models (VLMs) and Sets of Marks (SOM), improving both action generation accuracy and grounding precision. Device control agents, on the other hand, focus exclusively on visual interaction - using VLMs to process screenshots across different platforms (macOS, Windows, Android, iOS) and executing actions through tools like PyAutoGUI or bash scripting.

Web agents leverage browser automation frameworks (e.g., Selenium, Playwright) for direct control, while device control agents depend on screen coordinates. This makes web agents more efficient for tasks like URL navigation, which can be executed in a single command (e.g., `page.goto(newUrl)'). A hierarchical multi-agent system, where device control agents delegate web-specific tasks to web agents, offers an optimal architecture. In this work, we focus specifically on web agents.

\subsection{Current Landscape: Web Agent Research and Frameworks}
The field of web agent research and development encompasses five distinct categories:\\
\emph{Evaluation Environments:}  Gym-compatible platforms and benchmarks for web agent include WebArena \cite{zhou2023webarena}, VisualWebArena \cite{koh2024visualwebarena}, WorkArena \cite{workarena2024}, MiniWoB \cite{shi2017world}, BrowserGym, and WebShop \cite{yao2022webshop}.\\
\emph{Dataset Development:} Comprehensive datasets for agent training and evaluation include Mind2Web \cite{gou2024navigating} and  WebLINX \cite{lu2024weblinx}.\\
\emph{Algorithmic Advancement:} Research focusing on enhanced agent capabilities includes (1) Search/Reflection/Memory augmentation: Search Agent \cite{koh2024tree}, Agent Workflow Memory \cite{wang2024agent}, SteP \cite{sodhi2024step}; (2) Advanced planning: LATS \cite{zhoulanguage}, Reflective MCTS \cite{yu2024improving}; (3) Training methodologies: Mind2Web \cite{gou2024navigating}, WebLlama \cite{lu2024weblinx}.\\
\emph{Research-Oriented Frameworks:} Recent works have introduced research-oriented frameworks such as SeeAct \cite{zhenggpt}, OpenWebAgent \cite{iong2024openwebagent}, WebPilot \cite{zhang2024webpilot}, and AutoWebGLM \cite{lai2024autowebglm}.\\
\emph{Production-Ready Frameworks:} Agent-E \cite{abuelsaad2024agent}, BrowserPilot, LaVague, and Sentient. However, cloud-based versions like MultiOn Playground\cite{multion2024playground} and Emergence Web Automation\cite{emergence2024web} are commercial applications that are not open-sourced.

Despite significant advances in web agent research \& development, a critical gap exists in the ecosystem: the absence of a production-ready solution that  minimal serverless back-end configuration, intuitive user and browser
interfaces, while maintaining extensibility for emerging research developments such as search agents and MCTS. The LiteWebAgent Open-Source Suite addresses this gap, offering a comprehensive solution for VLM-based web agent applications.

\subsection{Key Contributions}
Key Contributions of the LiteWebAgent Open-Source Suite are as follows:

\begin{enumerate}
    \item \textbf{LiteWebAgent Agent Framework:} An extensible web agent framework that decouples action generation from action grounding. It supports various types of web agents, such as \textbf{FunctionCallingAgents} and \textbf{PromptAgents}. Agent planning and memory are integrated into \textbf{FunctionCallingAgents} to enhance long-horizon task planning, tackle complex workflows, and provide self-correction mechanisms for improved robustness during execution.

    \item \textbf{Agent Tree Search:} Extends LiteWebAgent to incorporate tree search, enabling exploration of multiple trajectories and balancing between exploitation and exploration.

    \item \textbf{Synchronous and Asynchronous APIs:} Seamlessly integrates with FastAPI for asynchronous calls and serverless functions, requiring minimal effort to deploy on platforms like Vercel for backend use.

    \item \textbf{Flexible User Interface:} Provides a system configuration panel and a comprehensive chat interface that features voice integration and task execution visualization.

    \item \textbf{Two Types of Deployed Systems:}
    \begin{enumerate}
        \item A production-ready Vercel-based web application that provides users with an agent-controlled remote browser.
        \item A Chrome extension that leverages LiteWebAgent's API to control an existing Chrome browser via the Chrome DevTools Protocol (CDP).
    \end{enumerate}
\end{enumerate}

\section{LiteWebAgent Agent Framework}
\begin{figure*}[t]
\includegraphics[width=\textwidth]{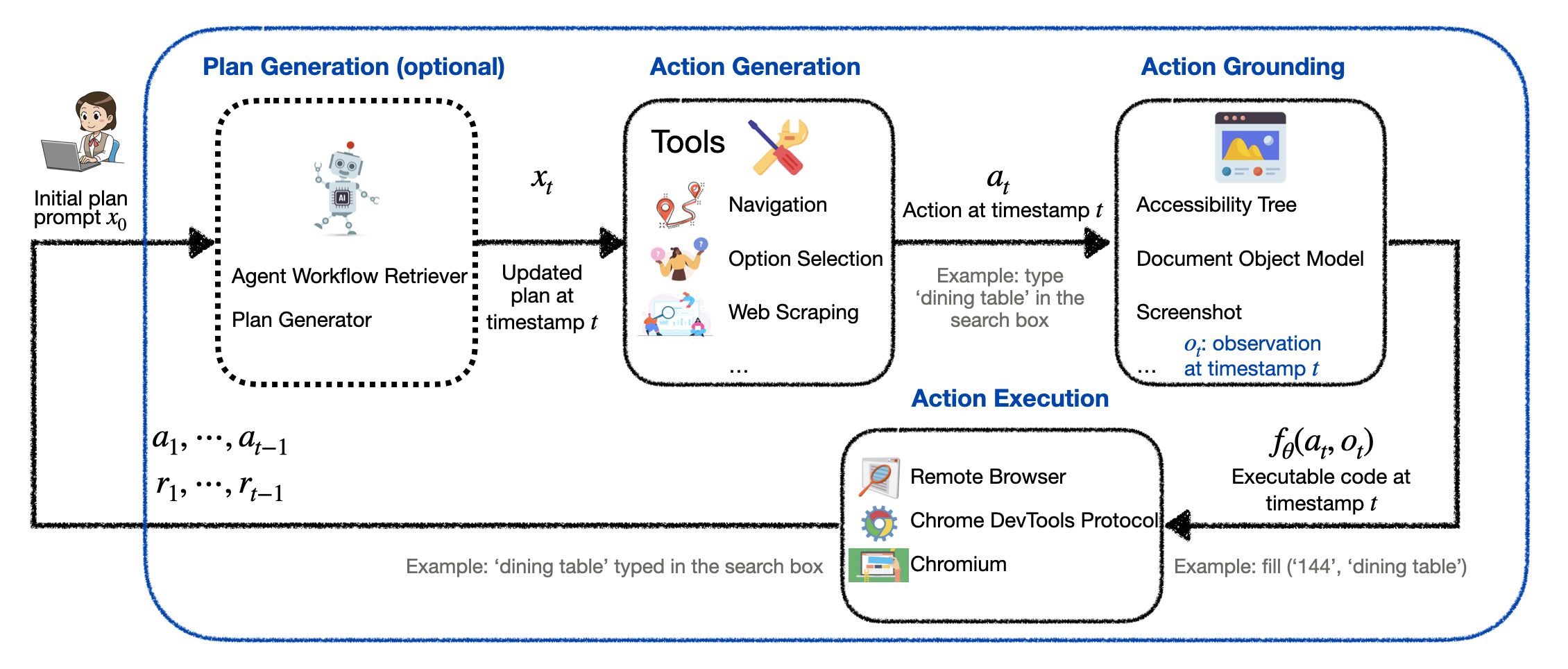}
\caption{Agent workflow}
\label{fig:agent-workflow}
\end{figure*}

In this section, we present the design of the web agent framework implemented in the open-source repository \textbf{LiteWebAgent}~\cite{zhang2024litewebagent}. Specifically, the framework decouples action generation from action grounding, supports multiple versions of agent planning, and integrates agent memory into the planning process. This design enables effective management of long-horizon tasks that require both planning and error recovery. Figure~\ref{fig:agent-workflow} illustrates an overview of the web agent's workflow. Moreover, we explain how the agent framework is extended with tree search to explore multiple trajectories and balance exploitation with exploration.

\subsection{Decouple Action Generation and Action Grounding}
We first define our problem.  We consider a contextual Markov decision process model represented by a tuple $\left(\mathcal{X}, \mathcal{S}, \mathcal{A}, \mathcal{M} \right)$, where $\mathcal{X}$ is the context space, $\mathcal{S}$ is the state space, and $\mathcal{A}$ is the action space. In our context, $x_0 \in \mathcal{X}$ is the initial plan prompt, which describes the plan of the task; $s_t\in \mathcal{S}$ is the complete state of the web environment; $a_t\in \mathcal{A}$ is the natural language description of the action at timestamp $t$. 
The mapping $\mathcal{M}$ maps a task prompt $x_0 \in \mathcal{X}$ to an episodic Markov Decision Process (MDP) model denoted as  $\mathcal{M}\left(x_t\right) = \left( \mathcal{S}, \mathcal{A}, \mathcal{T}, o_t, r_t \right)$ with  $o_t\in \mathcal{O}$ is the observation at time $t$; $r_t\in \mathcal{R}$ is the reward/evaluation at time $t$; and $T(s_t,a_t)\in \mathcal{T}$ is the state transition function. Examples of webpage observations include the Accessibility Tree, simplified DOM (Document Object Model), screenshot, Sets of Marks (SOM), and interactive elements.

At time step $t$, given the trajectory of rewards and actions $(r_{1}, \ldots, r_{t-1}, a_1, \ldots, a_{t-1})$, our agent generates the next action $a_t$ in natural language by considering the initial plan prompt $x_0$. That is, the VLM-based policy for action generation is defined as $a_t\sim\pi( x_0, r_1, \ldots, r_{t-1}, a_1, \ldots, a_{t-1})\in \Pi$, where $\Pi$ is the policy class. Note that 
 we cannot use $o_t$ directly since $o_t$ typically contains extensive semantic information, making it too long and consuming too many tokens, so we only pass the trajectory of $r_t$ as input to the VLM-based policy for action generation.

For the policy $\pi$, actions are selected using greedy decoding, where the highest-ranked action is chosen. Consider a pretrained language model $f_\theta(x)$, parameterized by $\theta$. When we use the model $f_\theta(x)$ with function calling, it can access web browser related tools like navigation for clicking elements, scrolling, and page navigation, as well as functionalities for uploading files, selecting options, and web scraping. With function calling, the policy is defined as $\pi = f_\theta^{\texttt{fc}}(x)$. Once no further function calls are triggered, the action generation process is considered complete, signaling that no additional actions are required. Alternatively, we can use few-shot prompting by providing some example actions in the prompt for the language model to generate the action, denoted by $\pi = f_\theta^{\texttt{prompt}}(x)$. However, unlike function calling, we will need to define a `FINISH' token explicitly in the context, indicating when the agent has completed its reasoning and action chain. In the LiteWebAgent framework, we set the first type of web agent as \textbf{FunctionCallingAgents} and the second type as \textbf{PromptAgents}.

After action generation, we use prompting for action grounding. Specifically, we employ a tailored prompting method that leverages the current webpage observation $o_t$ as context to transform the natural language action $a_t$ into executable actions, as a wrapper of Playwright code. This process can be formalized as $f_\theta^{\texttt{ag}}(a_t, o_t)$. Our framework is flexible, allowing different combinations of features to serve as the environment observation $o_t$. Examples include the Accessibility Tree (AXTree), simplified DOM (Document Object Model), screenshots, Sets of Marks (SOM), and interactive elements.

This decoupled approach provides greater flexibility and precision in controlling the web interaction process while significantly reducing the number of prompt tokens required for action generation. Especially by separating the two steps and adopting the \textbf{FunctionCallingAgents}, developers can seamlessly integrate new tools, such as web scraping or file read/write functions, enabling the web agent to have a broader range of capabilities within and beyond web browsing, and making it compatible for extension to multi-agent settings. In contrast, other web agent frameworks, which are primarily designed for browser interaction, lack this level of flexibility.

\subsection{Agent Planning}

In our framework, we distinguish between two concepts: goal and plan. The user is required to specify a goal, but the plan can be left empty. If no plan is provided, we generate an initial plan using a prompting method, based on the specified goal, represented as $x_0 = f_\theta^{\text{plan}}(\text{goal})$.

We employ different types of planning for \textbf{FunctionCallingAgents} agents in our framework:

\underline{Basic Function Calling Agent:} This agent leverages the LLM's planning ability to generate and execute function calls recursively, stopping when no further function calls are triggered. This simple stopping mechanism aligns with industry best practices, like those in Claude's computer use demos. This simple yet effective approach works surprisingly well for most tasks.

\underline{High-Level Planning Agent:} This agent replans based on action execution trajectory. The updated plan at time $t$ is generated as $x_t = f_\theta^{\text{plan}}(x_0, r_1, \ldots, r_{t-1}, a_1, \ldots, a_{t-1})$, and this plan is then used to guide the action generation process, $\pi(x_0, x_t, r_1, \ldots, r_{t-1}, a_1, \ldots, a_{t-1})$.

\underline{Context-Aware High-Level Planning Agent:} In addition to replanning based on action execution history, this agent incorporates context-aware information, such as the current environment observation (e.g., screenshots, Accessibility Tree). The updated plan at time $t$ is generated as $x_t = f_\theta^{\text{plan}}(x_0, o_t, r_1, \ldots, r_{t-1}, a_1, \ldots, a_{t-1})$, where $o_t$ represents the current observation. This allows for more informed decision-making based on the current context.

\subsection{Agent memory}

We incorporate Agent Workflow Memory (AWM) \cite{wang2024agent} into both the initial plan generation and replanning steps of the LiteWebAgent backend. In the initial plan generation, the process is defined as $x_0 = f_\theta^{\texttt{plan}}(\texttt{goal}, \texttt{AWM})$, where AWM, as relevant workflows to the current task, is used to inform the plan. During replanning, the updated plan at time $t$ is generated as $x_t = f_\theta^{\texttt{plan}}(x_0, s_t, \texttt{AWM}, r_1, \ldots, r_{t-1}, a_1, \ldots, a_{t-1})$, where the current state, as well as all previous evaluations and actions are taken into account.

\subsection{Agent Tree Search}

We extend \textbf{LiteWebAgent} with tree search capabilities, making it the first non-research framework to integrate a VLM-based web agent with tree search. The official repository will be available at: \url{https://github.com/PathOnAI/VisualTreeSearch-Demo}.

\subsubsection{Algorithm Implementations}

Rather than decoding a single trajectory, we explore multiple trajectories by sampling $k$ actions from the policy $\pi(x_0, r_1, \ldots, r_{t-1}, a_1, \ldots, a_{t-1})$, where $k$ is the branching factor. Each node $i$ in the search tree is associated with an action $a_t^i$ and its evaluation $r_t^i$. Backtracking through parent nodes reconstructs the entire trajectory.

We implement Breadth-First Search (BFS) and Depth-First Search (DFS) as \textbf{SimpleSearchAgent}. For more sophisticated exploration, \textbf{Monte Carlo Tree Search (MCTS)} balances exploitation and exploration using four steps:

\begin{enumerate}
    \item \textbf{Selection:} Starting from the root, select child nodes based on the Upper Confidence Bounds, balancing node value and visit count.
    \item \textbf{Expansion:} Expand a leaf node by sampling actions from $\pi$ and adding new child nodes.
    \item \textbf{Evaluation:} Use a VLM-based value function to score each trajectory with a scalar in $[0, 1]$.
    \item \textbf{Backpropagation:} Propagate the reward back to update node statistics along the path to the root.
\end{enumerate}

MCTS dynamically prioritizes promising nodes, improving decision efficiency. MCTS and variants are implemented as separate \textbf{SearchAgents}.

\subsubsection{Implementation Details}

\paragraph{Replay Module}  
We implement a robust replay module that executes action trajectories starting from an initial URL. Our implementation converts actions to Playwright code using unique selectors for precise element targeting. These selectors are generated by analyzing multiple element attributes - including ID, name, role, and tag name - while avoiding framework-specific patterns and non-deterministic IDs. For improved reliability, our implementation incorporates the element's position among siblings and relevant class names when needed to ensure uniqueness and accuracy."

For example, the unique selector for the Google search button is:

\noindent
\begin{minipage}{0.5\textwidth}
\begin{lstlisting}[language=, basicstyle=\ttfamily\small]
[role="search"] > div:nth-of-type(1) > 
div:nth-of-type(1) > div:nth-of-type(2) > 
div:nth-of-type(4) > div:nth-of-type(6) > 
center > input[type="submit"]
[aria-label="Google Search"]:nth-of-type(1)
\end{lstlisting}
\end{minipage}

This selector can be used in Playwright to locate and interact with the element.

\paragraph{VLM-Based Functions}  
We reuse VLM-based policy and reward functions from LiteWebAgent. A VLM-based value function evaluates trajectories $V(x_0, r_1, \ldots, r_{t-1}, a_1, \ldots, a_{t-1})$ through prompting.
\section{Demonstration}
\subsection{High level overview}

\begin{figure*}[t]
\includegraphics[width=\textwidth]{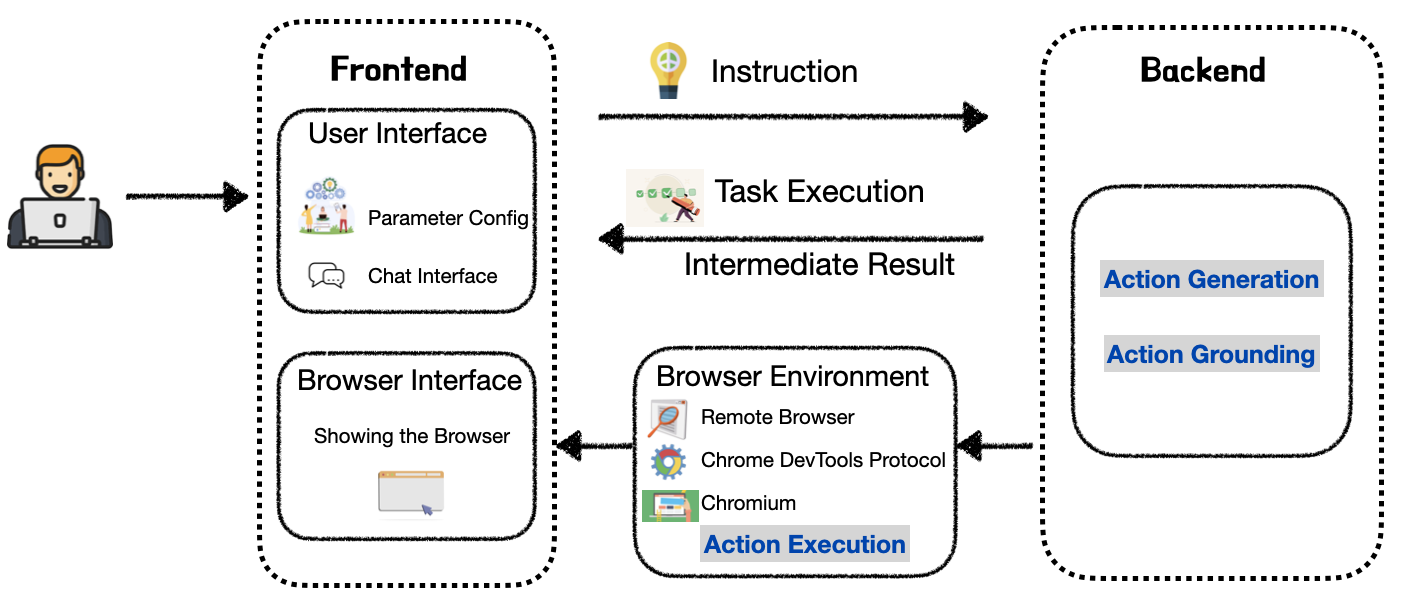}
\caption{System Design: High Level Overview}
\label{fig:system-high-level}
\end{figure*}

Figure~\ref{fig:system-high-level} provides a high-level overview of our implemented systems. The frontend includes a chat interface for user interaction and a browser interface that displays the web agent performing actions in the browser environment. Using the configuration panel in the user interface, users can set system parameters and send instructions to the backend via the chat interface. Our system supports three types of browser environments: (1) initializing a new Chromium browser, (2) controlling an existing Chrome browser via the Chrome DevTools Protocol (CDP), and (3) connecting to remote browsers.

The backend handles (1) initializing and connecting to the browser environment, (2) setting up the web agent, and (3) processing user instructions through action generation and grounding steps to produce executable Playwright wrapper code. This code is then executed in the browser environment, while the interface provides real-time visualization of agent actions.

While the agent performs actions, the backend also sends intermediate task execution results to the frontend, which are processed and displayed as intermediate steps on the frontend.

Sections~\ref{subsec:litewebagent-fullstack} and~\ref{subsec:litewebagent-chrome-extension} provide detailed explanations of the two fully deployed systems based on this design: a production-ready full-stack web application and a Chrome extension for controlling an existing Chrome browser.

\subsection{LiteWebAgent Full-Stack Deployed System}
\label{subsec:litewebagent-fullstack}

Several industry implementations allow remote browser control via iframes embedded in the frontend, enabling users to observe and interact with web agent actions (e.g., MultiOn Playground). Our project offers an open-source alternative, with frontend and backend endpoints deployed on Vercel.

\subsubsection{Backend}
Initially, we used BrowserGym \cite{workarena2024} with Playwright's synchronous API for the backend. Compatibility issues with FastAPI and asynchronous serverless functions led us to refactor the system to Playwright's asynchronous API. We integrated BrowserBase for remote browser sessions, leveraging the `session\_id’ to reconnect to active sessions and displaying the live browser URL in the frontend interface. These changes allowed us to deploy the backend as a serverless function on Vercel.

\subsubsection{Frontend}
Our frontend allows developers to create web agents and enables users to interact with them directly in the browser.

\begin{figure}
\includegraphics[width=0.5\textwidth]{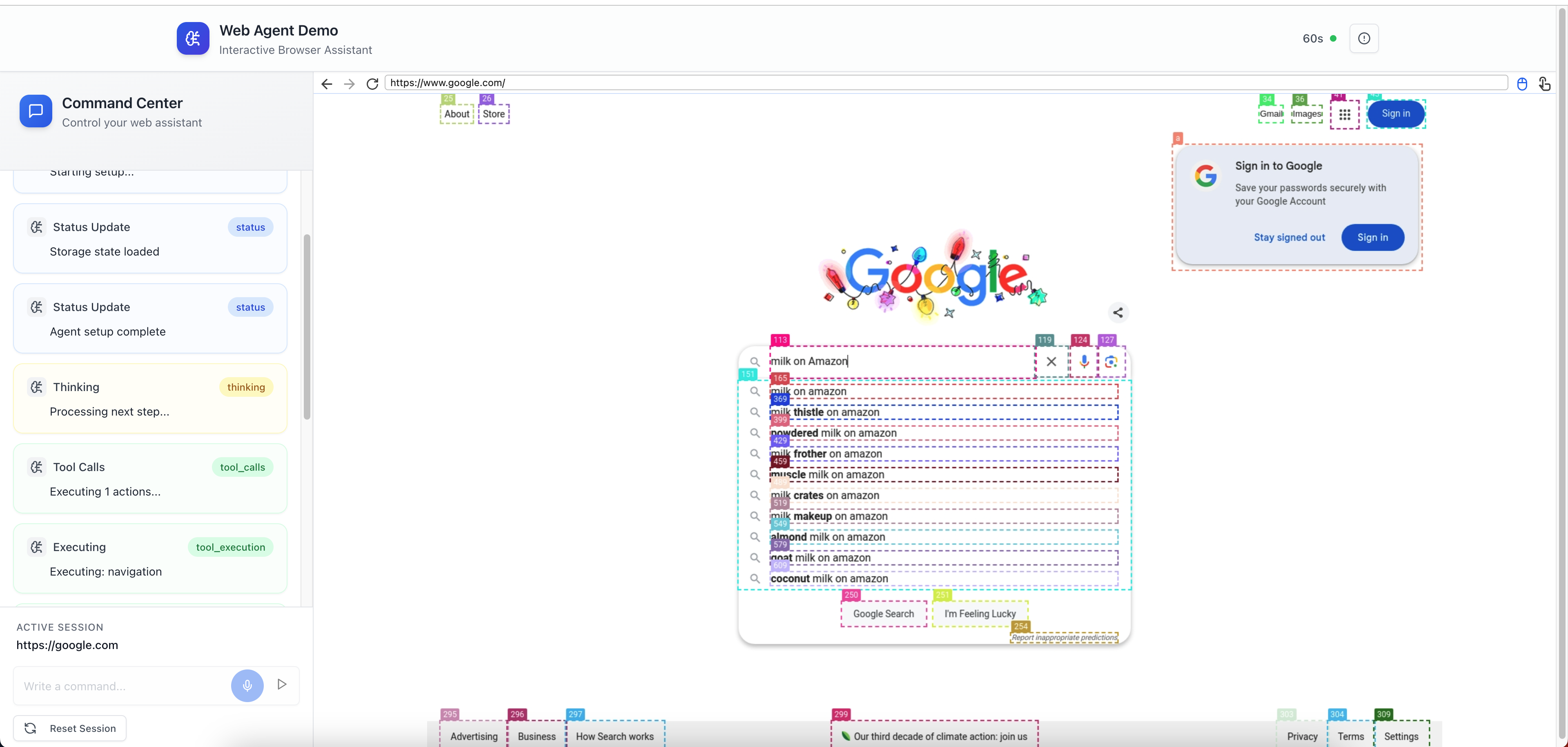}
\caption{Screenshot of frontend UI}
\label{fig:frontend-screenshot}
\end{figure}

\paragraph{User Interface}
Users interact with the system through a configuration panel and chat interface. They first set a starting URL and task goal, then sequentially send plan prompts via the chat interface. As the agent performs actions, intermediate results—including execution status, step-by-step action generation, action grounding output, and action execution results—are displayed in the frontend (Figure~\ref{fig:frontend-screenshot}).Additionally, the chat interface supports voice integration with TTS (Text-to-Speech) and ASR (Automatic Speech Recognition) powered by Deepgram.

\paragraph{Browser Interface}
Our system leverages BrowserBase for the remote browser environment. The frontend uses the session ID to retrieve the live browser URL and embeds it in an iframe for seamless interaction. To enhance the browser interface, we introduced additional demo effects: (1) interactive elements are highlighted as a Set of Marks to improve action grounding by VLMs (as shown in Figure~\ref{fig:frontend-screenshot}), and (2) before each action, relevant elements are highlighted with explanatory text boxes containing natural language descriptions.

\subsection{LiteWebAgent Chrome Extension}
\label{subsec:litewebagent-chrome-extension}

In addition to the web interface, we provide a Chrome extension that enables users to control local browser sessions by attaching Playwright to an existing browser instance. This approach allows LiteWebAgent to operate within a personalized browser context (e.g., sign-in status) and offers better privacy guarantees. The backend setup mirrors that of Section~\ref{subsec:litewebagent-fullstack}, but uses CDP (Chrome DevTools Protocol) instead of a remote browser environment.

\begin{figure}
\includegraphics[width=0.5\textwidth]{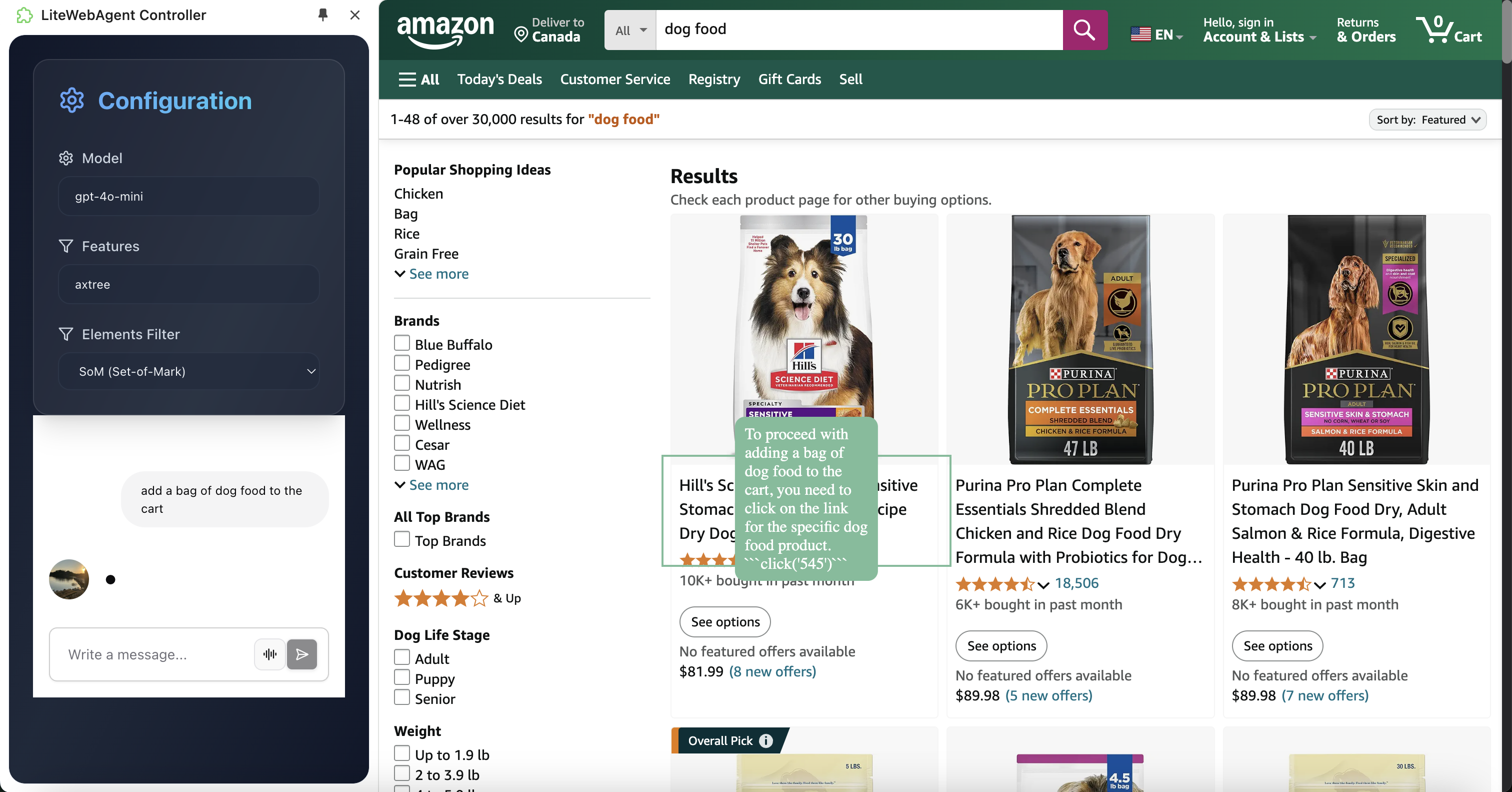}
\caption{Screenshot of Chrome extension UI}
\label{fig:chrome-extension-screenshot}
\end{figure}

The frontend interface is similar to the web interface. The extension appears in a side panel alongside the main webpage content. The configuration panel allows users to select models, choose action grounding features, and apply element filters to reduce token usage. Users can send prompts via the chat interface to instruct the agent to perform actions in the CDP browser environment. The browser interface displays agent actions alongside the target webpage, as shown in Figure~\ref{fig:chrome-extension-screenshot}.

\section{Conclusion and Future Work}  
\textbf{LiteWebAgent: The Open-Source Suite for VLM-Based Web-Agent Applications} offers a robust toolkit to address key gaps in the web agent ecosystem. It features: (1) an extensible core agent framework with simple yet effective baseline implementations, offering a scalable foundation for integrating new research such as agent workflows, memory management, planning, and tree search, while making it easy for developers to add additional tools beyond web browsing; (2) asynchronous APIs that integrate seamlessly with FastAPI, supporting serverless deployment on platforms like Vercel; (3) a flexible user interface with configuration options, a chat interface, voice integration, and task execution visualization; and (4) two types of browser environments correspond to the deployed systems: interactions with existing browsers via the Chrome DevTools Protocol (CDP) and a production-ready, Vercel-based full-stack application — an open-source, MultiOn-style service using a remote browser.

In the future, our objectives include: (1) extending the demo to build the first production-ready web agent with tree search demo with a user-friendly interface and intuitive visualization, (2) integrating LiteWebAgent into multi-agent frameworks such as \citet{zhang2024litemultiagent} to enable multi-agent frameworks with web browsing capabilities, (3) adding an evaluation module including new metrics to more comprehensively evaluate web agent performance.

\bibliography{custom}
\appendix

\end{document}